\documentclass[lettersize,journal]{IEEEtran}
\usepackage{amsmath,amsfonts}
\usepackage{algorithmic}
\usepackage{algorithm}
\usepackage{array}
\usepackage[caption=false,font=normalsize,labelfont=sf,textfont=sf]{subfig}
\usepackage{textcomp}
\usepackage{stfloats}
\usepackage{url}
\usepackage{verbatim}
\usepackage{graphicx}
\usepackage{cite}
\usepackage{hyperref}
\hypersetup{
}
\usepackage{makecell}
\usepackage{multirow}
\usepackage{amsmath}
\begin{document}

\title{VPRef: A Cross-Domain Benchmark for Referring Remote Sensing Image Segmentation}

\author{
\thanks{This work was supported in part by the Australian Government through the Australian Research Council's Discovery Projects Funding Scheme under Project DP220101634. (Corresponding author: Tao Huang.)} 
%
Quanwei Liu,~\IEEEmembership{Student Member,~IEEE}\thanks{Quanwei Liu is with the College of Science and Engineering, James Cook University, Cairns QLD 4878, Australia (e-mail: quanwei.liu@my.jcu.edu.au).}, 
Tao Huang,~\IEEEmembership{Senior Member,~IEEE}\thanks{Tao Huang is with the College of Science and Engineering, James Cook University, Cairns QLD 4878, Australia and the Centre for AI and Data Science Innovation, James Cook University, Cairns QLD 4878, Australia (e-mail: tao.huang1@jcu.edu.au).}, 
Jiaqi Yang,~\IEEEmembership{Member, ~IEEE}\thanks{Jiaqi Yang is with the Department of Forest and Wildlife Ecology, University of Wisconsin-Madison, Madison, WI 53706 USA (e-mail: jiaqi.yang@wisc.edu).},
Wei~Xiang,~\IEEEmembership{Senior Member,~IEEE}\thanks{Wei Xiang is with the School of Computing, Engineering and Mathematical Sciences, La Trobe University, Melbourne, VIC 3086, Australia (e-mail: W.Xiang@latrobe.edu.au).}
}

\markboth{Preprint. Under review.}%
{Shell \MakeLowercase{\textit{Liu et al.}}: VPRef}



\maketitle

\begin{abstract}
Rapid advancements in vision-language models have propelled Referring Remote Sensing Image Segmentation (RRSIS) to the forefront of Earth observation. 
However, practical deployments suffer severe performance degradation under a coupled dual-drift paradigm: visual domain drift from cross-spatial-resolution mismatches and spectral variations, alongside textual drift from unconstrained, variable user-input granularities. 
To mitigate these bottlenecks, this paper establishes the first cross-domain RRSIS benchmark, designated as the Vaihingen-Potsdam Referring (VPRef) dataset, comprising 39\,789 language-image-annotation triplets organized into a three-tier linguistic hierarchy. 
Building upon this benchmark, we develop a tailored parameter-efficient domain adaptation baseline anchored on the Segment Anything Model 3 (SAM3) via Low-Rank Adaptation (LoRA). 
Our framework counteracts visual distribution discrepancies through pseudo-label-driven self-training and addresses textual drift via random multi-granularity text prompt mixing. 
Crucially, the distribution of empirical metrics across ablative variants suggests that the two adaptation strategies affect metrics differently: cross-modal semantic robustification drives fine-grained semantic invariance, while pseudo-label propagation governs macro-scale spatial grid alignment. 
Extensive benchmarks demonstrate that the proposed framework achieves competitive cross-domain segmentation boundaries while modifying merely 1.09\% of the foundational parameter footprint, establishing a robust baseline for future multi-modal remote sensing domain adaptation research.
The dataset and code will be available at \url{https://github.com/quanweiliu/VPRef}.
\end{abstract}


\section{Introduction}
\IEEEPARstart{R}{apid} advancements in high-resolution remote sensing (HSR) imagery and the rise of multimodal vision-language models have propelled Referring Remote Sensing Image Segmentation (RRSIS) to the forefront of Earth observation research \cite{li2024model, liu2025pixels}. 
Compared with traditional semantic segmentation or object detection frameworks that are restricted to predefined, closed-set categories, RRSIS offers an inherently user-friendly interaction paradigm. 
In practical deployment scenarios, such as disaster response \cite{yuan2025effc}, urban planning \cite{liu2026knowcl} and smart agriculture \cite{dong2022weighted}, non-expert operators can bypass tedious professional labels and manual boundary tracing. 
Instead, by providing intuitive natural language instructions, such as ``\textit{the blue-roofed factory building on the north side of the road and partially occluded by trees}", users enable the model to achieve pixel-level localization within large-scale imagery \cite{yuan2024rrsis}. 
This natural-language-as-an-interface paradigm substantially unlocks the practical utility of remote sensing big data.

Transitioning RRSIS models from controlled laboratory environments to unconstrained real-world applications exposes severe generalization vulnerabilities \cite{wang2021loveda}. 
The primary bottleneck stems from inherent visual domain drift across remote sensing datasets.
Current state-of-the-art architectures, including LAVT \cite{yang2022lavt} and RMSIN \cite{liu2024rotated}, are evaluated almost exclusively within a closed, single-domain setting where the training and testing distributions share identical geographic regions and sensor configurations. 
Actual deployments, however, require models to generalize across heterogeneous cities and variable acquisition conditions \cite{wang2025nl2net, wang2024directnet}. 
Pronounced variations in regional landscapes, sensor spectral characteristics, and particularly the non-linear scaling induced by cross-spatial resolution mismatches, such as migrating from a 9 cm ground sampling distance (GSD) to a 5 cm GSD, disrupt the learned cross-modal semantic alignment, leading to catastrophic performance degradation on unseen target domains \cite{zhang2021curriculum}.

A more insidious challenge arises from the unconstrained, non-standard nature of real-world user queries. 
Human linguistic habits exhibit substantial variation. Some users favor brief, macro-level descriptions restricted to basic categories and scales. 
For example, the client might use ``the small cars" to find all cars scattered across a remote sensing image.
In contrast, other users want to articulate highly intricate sentences with multiple topological relations to disambiguate dense clusters of similar objects. 
This severe fluctuation in linguistic granularity triggers drift in text across domains. 
%
%
Existing RRSIS benchmarks fail to capture this vulnerability because they rely on rigid, template-based generation or provide only a single textual level, leaving models ill-equipped to handle the dual drift in visual and linguistic features \cite{liu2024rotated, dong2024cross, yang2025large}.

To address these limitations, we introduce the first cross-domain RRSIS benchmark, the Vaihingen-Potsdam Referring (VPRef) dataset. 
Transcending single-domain constraints, VPRef bridges two classic remote sensing domains, ISPRS Vaihingen and ISPRS Potsdam \cite{rottensteiner2014results}, which exhibit distinct spectral profiles and a pronounced shift in spatial resolution. 
The dataset comprises 8763 high-resolution image patches paired with 39\,789 high-quality language annotations. To replicate authentic, non-standard user inputs, VPRef replaces monotonous template configurations with a metadata-driven pipeline assisted by advanced large language models (LLMs). 
This approach yields a three-tier hierarchy of linguistic granularity: Basic, Moderate, and Complex. The Basic tier omits relative positions and focuses on object attributes and scale; the Moderate tier introduces first-order spatial relationships; the Complex tier integrates multi-layered geometric constraints, target occlusion states, and chained topological reasoning. 
As analyzed in Table \ref{vpdataset}, VPRef provides a rigorous diagnostic tool to evaluate model resilience against multi-dimensional domain shifts.


Utilizing the VPRef benchmark, we execute an extensive empirical evaluation across prominent RRSIS architectures, revealing a performance gap of approximately 20 percentage points under spatial-domain transitions and 47.88 percentage points under textual-domain transitions.
To address this performance drop, we present a tailored domain adaptation baseline designed to counteract both visual and textual shifts. 
For visual domain drift, we first introduce and evaluate a strong semantic baseline based on a LoRA-adapted Segment Anything Model 3 (SAM3).
Then, we integrate a self-training mechanism that propagates high-confidence pseudo-labels from the target domain back into the source training loop. 
To address textual domain drift, we concatenate multi-granularity text prompts and use a random sampling strategy during training, simulating variance in heterogeneous user inputs. 
Experimental results show that these joint strategies effectively mitigate dual drift, establishing a robust baseline for future cross-domain RRSIS research.


In summary, the primary contributions of this work are threefold:

\begin{enumerate}
\item{We establish the VPRef dataset, the first cross-domain benchmark for RRSIS. The dataset provides 39\,789 high-quality language-image-annotation triplets structured across a three-tier linguistic hierarchy to accurately reflect unconstrained user commands.}
\item{We develop a domain adaptation baseline to mitigate visual and textual drift. The framework counteracts visual drift through self-training and addresses textual drift via random sampling from concatenated, multi-granularity text prompts.}
\item{We provide the first systematic analysis showing metric-dependent effects of the two adaptation strategies and highlight limitations of current models under cross-domain scenarios.}
\end{enumerate}

The remainder of this paper is organized as follows. 
Section \ref{relatedwork} reviews related work on referring image segmentation datasets and methods. 
Section \ref{VPRefData} details the construction, properties, and hierarchical annotation pipeline of the proposed VPRef dataset. 
Section \ref{methodology} describes the proposed domain adaptation framework, detailing the self-training mechanism and the multi-granularity text sampling strategy. 
Section \ref{experiments} presents the experimental setup and quantitative and qualitative benchmark results. 
Finally, Section \ref{conclusion} concludes this paper.



\section{Related Work}
\label{relatedwork}

\subsection{Referring Image Segmentation} 
\label{subsec:ris_review} 

Referring image segmentation originated in the natural image domain, where the objective is to segment a specific target given an arbitrary natural language expression. 
Early paradigms established benchmarks such as RefCOCO, RefCOCO+ \cite{kazemzadeh-etal-2014-referitgame}, and RefCOCOg \cite{Mao_2016_CVPR}, utilizing recurrent neural networks \cite{Li_2018_CVPR} or cross-attention mechanisms \cite{hu2020bi, liu2021cross} to fuse visual and textual features. 
The field shifted significantly with the advent of the Vision Transformer (ViT). For instance, Language-Aware Vision Transformer (LAVT) \cite{yang2022lavt} showed that early-stage multi-modal fusion within the vision backbone yields superior cross-modal alignment compared to late-fusion architectures. 
Other developments, including scale-aware attention and query-based detectors \cite{Liu_2023_CVPR}, further enhanced the suppression of linguistic noise and improved boundary localization.

Transitioning this paradigm to Earth observation led to the development of RRSIS. Remote sensing imagery introduces unique challenges, including overhead perspectives, extreme scale variations, and dense target distributions. 
To capture these characteristics, early benchmarks such as RefSegRS \cite{yuan2024rrsis} and RRSIS-D \cite{liu2024rotated} introduced template-based and manually written textual descriptions for high-resolution imagery. 
To handle the arbitrary orientations of geographical features, architectures such as the Rotated Multi-Scale Interaction Network (RMSIN) \cite{liu2024rotated} incorporate oriented bounding mechanisms and multi-scale attention to align text with rotated targets. 
More recent large-scale benchmarks, such as RISBench \cite{dong2024cross} and NWPU-Refer \cite{yang2025large}, expand the diversity of geographic scenes and feature granularities. 
However, existing RRSIS frameworks assume that training and testing datasets come from the same domain and use consistent referring expressions. 
Consequently, these models lack the mechanisms required to handle structural and spectral discrepancies when deployed across heterogeneous geographic regions.

\subsection{Domain Adaptation in Remote Sensing Imagery} 
\label{subsec:da_review} 

Domain Adaptation (DA) addresses the performance degradation caused by domain shift, which occurs when a model trained on a source dataset is then deployed on an unannotated target dataset \cite{li2024comprehensive}. 
In remote sensing semantic segmentation, domain shifts arise from variations in sensor specifications, illumination conditions, atmospheric effects, and regional architectural styles \cite{peng2022domain}.
Early remote sensing DA methods relied heavily on adversarial learning to align global feature distributions between domains, either at the input pixel level or within intermediate latent spaces \cite{tasar2020standardgan, wang2023ftsgan}.
Albeit effective for coarse style transfer, generic domain-minimax games frequently disrupt fine-grained class boundaries and exhibit volatile gradient optimization when coupled with foundation models \cite{tsai2018learning, zou2018unsupervised}. 
Consequently, the research paradigm has fundamentally transitioned toward target-domain pseudo-annotation mining via iterative self-training loops.

To counteract pseudo-label noise under severe domain shifts, recent self-training frameworks incorporate advanced distribution alignment and denoising mechanisms.
For instance, Liang et al. \cite{liang2025spectral} utilize probability-consistent self-training paired with batch nuclear-norm maximization to align features while preserving prediction diversity. 
Similarly, Liang et al. \cite{liang2023unsupervised} integrate adversarial learning with self-training, leveraging a conditional adversarial loss to filter target-domain prediction noise and stabilize classifier adaptation.
These methods iteratively select high-confidence model predictions on unannotated target imagery to optimize the network via supervised objectives. 
To mitigate the accumulation of confirmation bias and label noise caused by spectral or resolution mismatches, researchers employ uncertainty estimation, class-balanced thresholds, and teacher-student consistency regularizations \cite{wang2021loveda, luo2024crots}.

In addition, the LoveDA \cite{wang2021loveda} dataset established a benchmark for evaluating these frameworks in remote sensing across distinct urban and rural landscapes, while ``Cross-city matters" \cite{hong2023cross} highlighted visual domain gaps across different metropolitan areas. 
Despite these advances in computer vision, conventional remote sensing DA methods and datasets remain limited to single-modality settings \cite{li2025enhanced}. 
They are unequipped to handle multi-modal RRSIS tasks, where visual domain drift couples with text domain drift to disrupt cross-modal semantic alignment.

\subsection{Efficiency and Parameter Scaling in Vision Models}
\label{subsec:efficiency_review}

The rapid scaling of vision-language foundation models has introduced a trade-off between reasoning capability and computational efficiency. 
Autoregressive vision-language models and modern reasoning segmentation frameworks leverage multi-billion-parameter backbones to interpret complex, implicit queries. 
Although these large-scale models exhibit remarkable zero-shot generalization, their parameter scaling introduces massive computational overhead \cite{quenum2026lisat, lai2024lisa}. 
This requirement creates a substantial barrier for remote sensing applications, which often demand rapid processing, onboard satellite execution, or deployment on edge devices with constrained memory and power resources \cite{gao2026pys}. 
Furthermore, executing unsupervised domain adaptation loops on models with billions of parameters is structurally unstable and computationally restrictive.
To balance performance with deployability, parameter-efficient fine-tuning (PEFT) and specialized foundational backbones have emerged as viable alternatives \cite{houlsby2019parameter}.

The SAM series demonstrates strong zero-shot localization capabilities by utilizing promptable visual features \cite{kirillov2023segment}. 
The recent SAM3 architecture integrates explicit concept representations, allowing the model to bridge low-level geometry with high-level semantic abstractions \cite{carion2026sam3segmentconcepts}. 
By utilizing parameter-efficient transfer learning techniques such as Low-Rank Adaptation (LoRA), researchers can adapt these foundation models to specific downstream tasks without updating the entire parameter space \cite{hu2022lora}. 
This study focuses on referring segmentation rather than generative-reasoning segmentation to ensure our framework remains highly parameter-efficient. 
This structural choice enables the integration of multi-tiered linguistic consistency constraints to stabilize cross-domain adaptation.

\begin{table*}[]
\centering
\caption{Comparative analysis of VPRef and prior RRSIS datasets.}
\label{vpdataset}
\begin{tabular}{lllllllll}
\hline
Dataset & Text Levels & Domains & GSD (m) & Images & Annotations & Resolution & Mask    Generation & Text    Generation \\
\hline
RefSegRS \cite{yuan2024rrsis} & 1 & 1 & 0.13m & 4420 & 4420 & 512 & Manual & Template \\
RRSIS-D \cite{liu2024rotated} & 1 & 1 & 0.5m–30m & 17\,402 & 17\,402 & 800 & Semi-auto & Template \\
RISBench \cite{dong2024cross} & 1 & 1 & 0.1m-30m & 52\,472 & 52\,472 & 512 & Semi-auto & Semi-auto(GPT-4V) \\
NWPU-Refer \cite{yang2025large} & 2 & 1 & 0.12m–0.5m & 15\,003 & 49\,745 & 1024-2048 & Manual & Manual \\
\textbf{VPRef (Ours)} & 3 & 2 & 0.05-0.09 & 8763 & 39\,789 & 480-864 & Manual & Semi-auto(Gemini 3.5) \\
\hline
\end{tabular}
\end{table*}

\begin{figure*}[!t]
\centering
\includegraphics[width=7in]{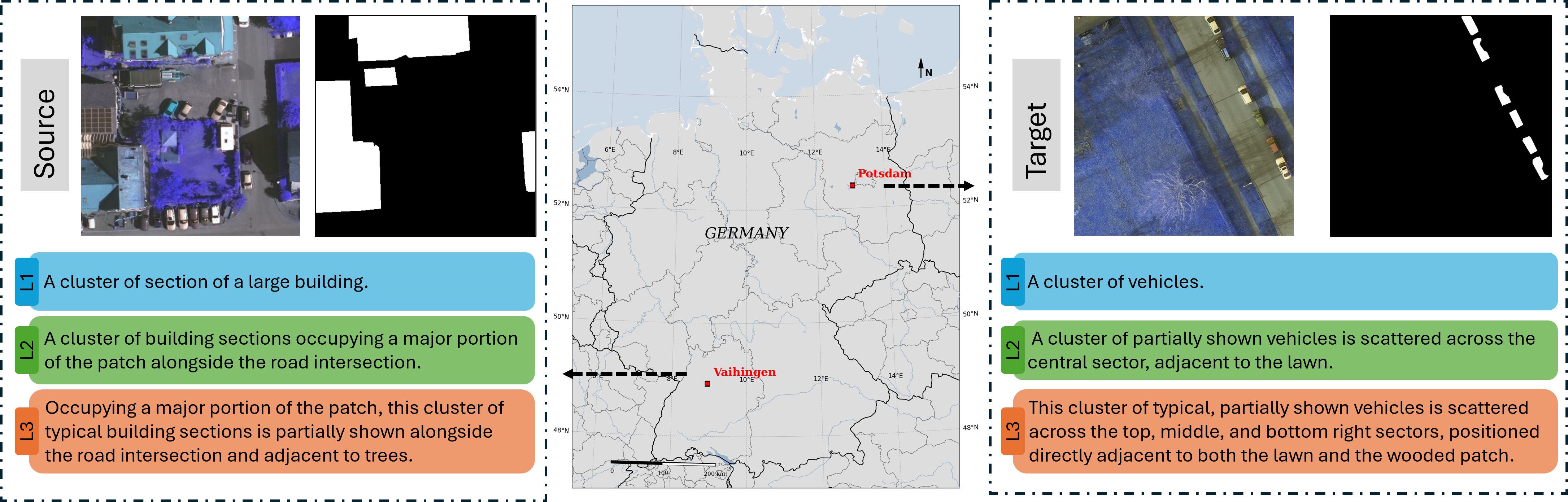}
\caption{Illustration of the VPRef dataset, which contains three levels: Basic, Moderate, and Complex. The L1, L2, and L3 denote the Basic, Moderate, and Complex tiers of linguistic description, respectively.}
\label{vpref}
\end{figure*}

\section{The VPRef Dataset}
\label{VPRefData}

To systematically investigate the dual drift of visual and textual features in Earth observation, we introduce the VPRef dataset. Transcending the single-domain boundaries of prior RRSIS benchmarks, VPRef provides a paired, multi-tiered linguistic evaluation platform spanning two distinct geographic domains. 
As summarized in Table \ref{vpdataset}, the dataset comprises 39\,789 high-quality language-image-annotation triplets, establishing a standardized baseline for cross-domain RRSIS research.

\subsection{Data Source and Domain Specifications}
We derive the visual components of VPRef from the classic ISPRS Vaihingen and ISPRS Potsdam aerial frameworks. 
Both domains utilize near-infrared, red, and green (IRRG) spectral band configurations. 
For the source domain, VaiRef, we extract 2510 image patches of size 480 × 480 pixels from the Vaihingen tiles, using the official semantic annotations to construct the reference masks. 
Following the tile assignments used in previous work \cite{hosseinpour2022cmgfnet}, the training tiles are 1, 3, 5, 7, 13, 17, 21, 23, 26, 32, and 37, while the validation tiles are 11, 15, 28, 30, and 34. The remaining tiles used in this study constitute the test set.
For the target domain, PotsRef, we extract 6253 image patches of size 864 × 864 pixels from the Potsdam tiles. The training tiles are $2\_10$, $3\_10$, $3\_11$, $3\_12$, $4\_11$, $4\_12$, $5\_10$, $5\_12$, $6\_8$, $6\_9$, $6\_10$, $6\_11$, $6\_12$, $7\_7$, $7\_9$, $7\_11$, and $7\_12$. The validation tiles are $2\_11$, $2\_12$, $4\_10$, $5\_11$, $6\_7$, and $7\_8$. The remaining tiles used in this study constitute the test set.
For both domains, we assign the original tiles to their respective partitions before patch extraction. We then extract patches with horizontal strides of 240 pixels and 432 pixels, respectively, corresponding to a 50\% overlap between adjacent patches. 
This overlap is restricted to patches within the same tile and partition. No original tile contributes patches to more than one partition, and the training, validation, and test sets contain no overlapping image patches. All referring expressions associated with a patch inherit the same partition.

Crucially, the two domains exhibit distinct geometric and spatial layout properties. VaiRef features a GSD of 9~cm, while PotsRef has a finer GSD of 5~cm.
This structural difference creates an explicit cross-spatial resolution mismatch, forcing models to adapt across different pixel scales, object-density profiles, and sensory footprints.

\begin{figure*}[!t]
\centering
\includegraphics[width=7in]{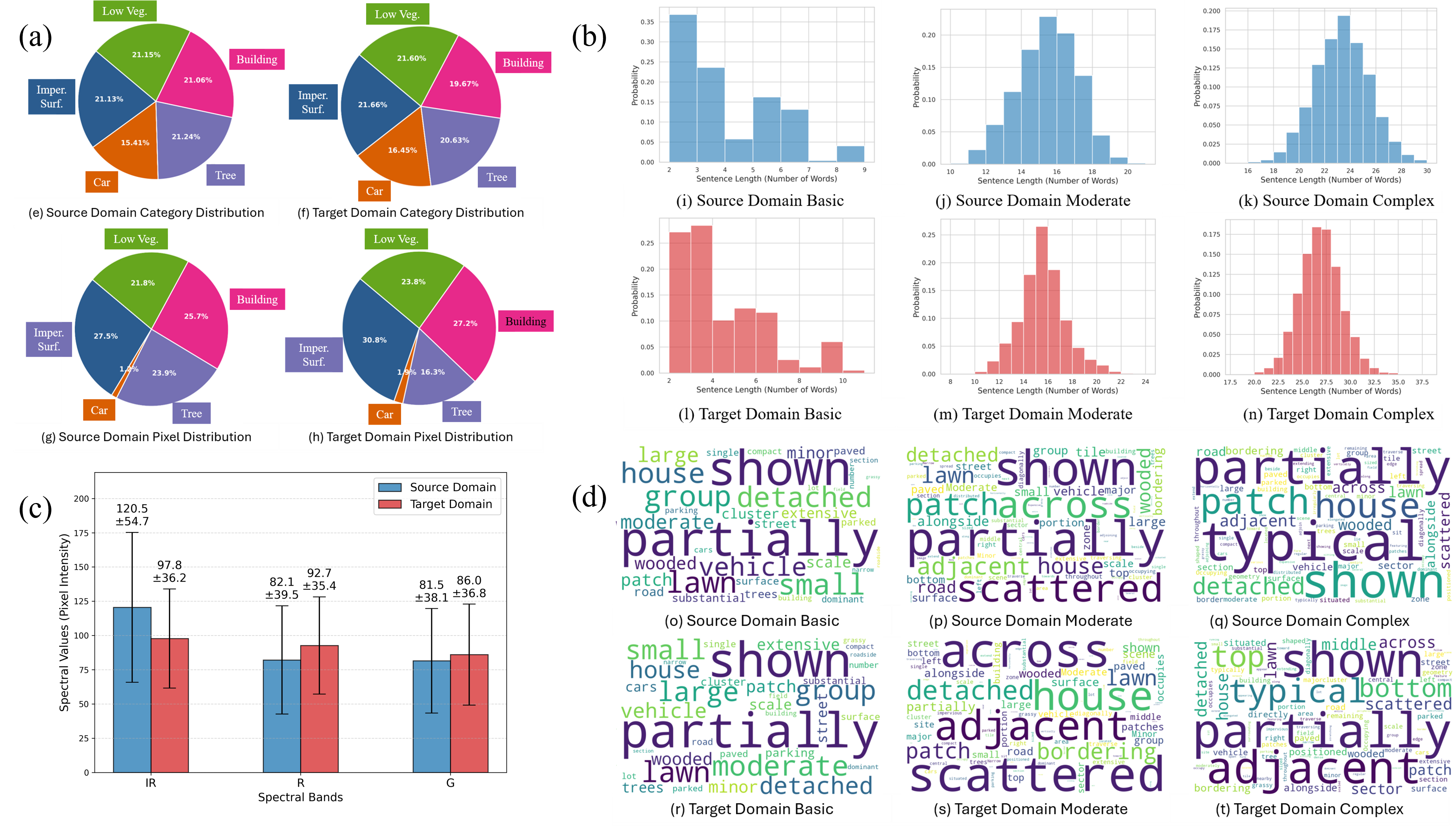}
\caption{Statistical analysis of the constructed VPRef dataset across the source and target domains. (a) Category and pixel proportion of five objects. (b) Distribution of word lengths of referring expressions. (c) Mean and standard deviation of spectral values. (d) Word clouds for referring expressions.}
\label{statistic}
\end{figure*}

\subsection{Metadata Extraction and Hierarchical Text Generation}
To build a high-quality multimodal dataset without manual text-writing fatigue or template monotony, we implement a metadata-driven pipeline powered by advanced generative models. 
The text generation process executes in three sequential stages: structural attribute mining, rule-based natural language mapping and text synthesis.
First, we develop a geometric extraction routine to parse the ground-truth semantic masks of each image patch. 
For each image patch and target semantic class, the reference mask is defined as the union of all pixels annotated with that class in the original semantic ground truth. 
This script isolates connected components for five target categories: impervious surfaces, buildings, cars, trees, and low vegetation. 
For each isolated instance or collective cluster, the routine computes explicit spatial attributes, including normalized bounding box coordinates, global centroids, aspect ratios, area ratios, and boundary truncation states. 
Because the underlying imagery relies on IRRG band configurations rather than standard RGB channels, we explicitly omit color attributes from the metadata extraction to prevent cross-modal spectral mapping confusion. 
Second, we design a rule-based mapping system that projects these spatial attributes into natural language, such as determining building scale and the relative positions of land objects.
The proposed semantic lifting pipeline decouples spatial masks from rigid numerical values, transforming shape and density metrics into granular, scale-invariant natural language primitives.
%
%
Third, the extracted structural metadata is passed to a specialized text generation script that interfaces with the Gemini 3.5. Rather than relying on rigid string concatenation, we implement a prompt-based synthesis mechanism to ensure grammatical fluidity and stylistic variation. 
The script processes macro attributes to construct basic descriptions. Concurrently, it ingests advanced spatial relations and chained topological structures to generate more sophisticated expressions. 
%

\subsection{Linguistic Granularity Tiers}
To replicate the diverse, unconstrained inputs of real-world end-users, VPRef establishes a three-tier hierarchy of linguistic granularity. Each tier represents a distinct level of complexity in cross-modal reasoning, as illustrated by the visual examples in Fig. \ref{vpref}.

\begin{itemize}
\item \textbf{Basic Tier:} Designed to simulate direct, low-overhead user inputs. This level completely excludes relative positioning or environmental context. It focuses exclusively on target scale and the core category name, yielding lean expressions such as ``a cluster of vehicles" or ``minor lawn".
\item \textbf{Moderate Tier:} Represents conventional operational queries. This tier incorporates first-order absolute locations and primary contextual relations. It describes targets using a direct subject-first syntax, typically constraining the description length to 12–20 words, such as ``a cluster of building sections occupying a major portion of the patch alongside the road intersection".
\item \textbf{Complex Tier:} Modelled after detailed, highly descriptive queries required for dense target disambiguation. This level integrates multiple chained topological relations, fine-grained shape geometries, and explicit boundary truncation states. The syntax varies dynamically across 22–32 words, forcing models to resolve multi-layered geometric constraints, such as ``occupying a major portion of the patch, this cluster of typical building sections is partially shown alongside the road intersection and adjacent to trees".
\end{itemize}

The Basic, Moderate, and Complex expressions are generated for this same mask. Additional attributes and spatial relations describe the collective target and its context without selecting a different subset of regions. 
In this study, textual drift refers to controlled variations in referring expressions across description levels, involving changes in descriptive granularity, linguistic complexity, and contextual relational content.

\subsection{Statistical Analysis and Dataset Characteristics}

Deep cross-domain statistics reveal the severe domain gaps embedded within VPRef, as visualized across the comprehensive analytics in Fig. \ref{statistic}. 
The spectral value distributions, illustrated in Fig. \ref{statistic}(c), show pronounced shifts in pixel intensities between the Vaihingen and Potsdam environments, driven by variable seasonal flight times and sensor calibrations. 
The word clouds presented in Fig. \ref{statistic}(d) illustrate the textual drift across the granularity scale. The Basic tier relies heavily on simple quantifiers, while the Moderate and Complex tiers introduce a dense vocabulary of spatial prepositions, geometric descriptors, and directional terms. 
This linguistic progression, coupled with the resolution shift from 9~cm to 5~cm, makes VPRef a highly diagnostic benchmark for quantifying model resilience to simultaneous visual and textual drift.

In addition, the image and language in the source and target domains show a consistent trend.
Figs. \ref{statistic}(e)-(f) highlight the land cover categories. The five objects are relatively evenly distributed in the dataset.
In contrast, Figs. \ref{statistic}(g)-(h) demonstrate that the Car accounts for only a very small proportion of pixels of the dataset.
This shows that, compared with other categories, the car is much smaller in scale, which may challenge the model's ability to recognise small objects.
Linguistically, the three tiers demonstrate clear structural separation. As depicted in Fig. \ref{statistic}(b), the sentence length of the Basic tier ranges from 2–10 words, the Moderate tier from 12–20 words, and the Complex tier from 22–32 words.


\begin{figure}[!t]
\centering
\includegraphics[width=3.5in]{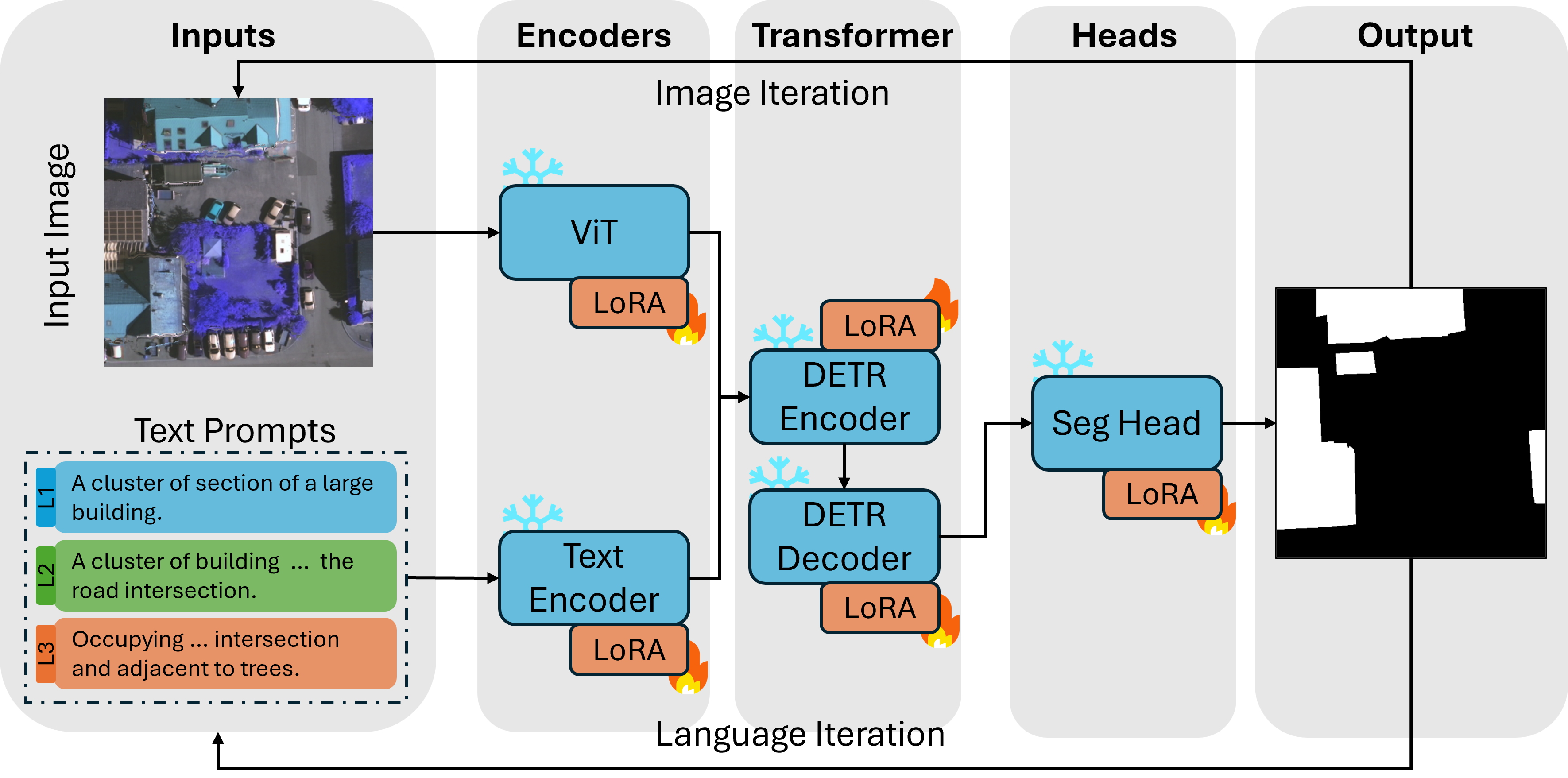}
\caption{Illustration of full LoRA Adaptation SAM 3 Architecture with self-training and multi-granularity text prompt mixing.}
\label{framework}
\end{figure}

\section{Methodology}
\label{methodology}

This section details a unified framework to counteract dual drift in visual appearance and referring expressions in cross-domain RRSIS. 
We first present the overall framework overview, followed by the PEFT mechanism based on SAM3-LoRA. 
We then describe the generalized domain adaptation components, comprising pseudo-label-driven self-training and a multi-granularity text prompt mixing strategy, which are extended symmetrically across all evaluated baseline architectures to establish a universal adaptation benchmark. 
Finally, we formulate the cross-modal matching logic and joint loss optimization objective.

\subsection{Overall Framework Overview}
The principal goal of our approach is to facilitate the cross-domain generalization of an RRSIS model from a labelled source domain to an unannotated target domain.
To address the simultaneous shifts in spatial-spectral visual features and in unconstrained user-command granularities, we establish a dual-drift adaptation pipeline, as illustrated in Fig. \ref{framework}.

The overall forward pass operates on a multi-modal interactive pipeline. Given an input remote sensing image patch paired with a natural language query, the framework extracts concurrent feature streams through a visual backbone and a textual encoder. 
These disparate modalities are projected into a shared embedding space and fed into a multi-modal interaction transformer, which coordinates fine-grained cross-modal attention layers to align linguistic tokens with pixel-level spatial representations. 
A specialized segmentation head then decodes these aligned tokens to output coordinate bounding boxes and dense pixel masks.

Rather than treating domain adaptation as an isolated vision task, our framework views the structural variance of text prompts as a regulatory mechanism. 
The self-training pipeline handles visual distribution alignment by propagating target domain predictions, while the text prompt mixing strategy stabilizes cross-modal text alignment. 
Crucially, these two adaptation components are designed as generalized modules. They operate independently of the underlying network architecture, enabling uniform deployment across conventional RRSIS baselines such as LAVT \cite{yang2022lavt} and RMSIN \cite{liu2024rotated}, as well as our advanced foundational baseline.

\subsection{Parameter-Efficient Fine-Tuning via SAM3-LoRA}
To establish a strong semantic baseline that handles complex Earth observation categories across domains, we anchor our main architecture on SAM3. 
Due to the massive scale of foundational parameters, full-weight backpropagation risks catastrophic forgetting and training divergence under unsupervised domain adaptation loops. 
We mitigate this instability by implementing LoRA, restricting gradient updates to a lightweight set of intrinsic rank-decomposition matrices while freezing the global foundational weights.

As illustrated in the architectural breakdown in Fig. \ref{framework}, we systematically insert LoRA layers into the network's core components to achieve balanced multimodal adaptation. 
Within the visual backbone, we embed LoRA layers into the query, key, and value projection weights of the vision transformer attention blocks to capture regional landscape variations. 
Symmetrically, the text encoder incorporates LoRA layers within its multi-head attention layers to handle domain-specific geographic terminology. 
To align these adapted feature streams, we embed low-rank matrices into both the encoder and decoder structures of the multi-modal interaction transformer, forcing the cross-modal queries to adapt to scale transitions. 
Finally, the prediction and segmentation heads are augmented with LoRA to refine boundary localization.

Formally, for any targeted weight layer $W_0 \in \mathbb{R}^{d \times k}$, the adapted forward pass bypasses full matrix modification by computing a parallel low-rank update. The modified weight matrix $W$ is expressed as:
\begin{equation}
W = W_0 + \Delta W = W_0 + \frac{\alpha}{r} (B \cdot A),
\end{equation}
where $B \in \mathbb{R}^{d \times r}$ and $A \in \mathbb{R}^{r \times k}$ represent the trainable low-rank matrices, $r \ll \min(d, k)$ denotes the intrinsic rank, and $\alpha$ is a constant scaling hyperparameter. During training, $W_0$ remains strictly frozen, and only $A$ and $B$ receive gradient updates. This design reduces the trainable parameter footprint to a fraction of the total model volume, ensuring computational efficiency during cross-domain adaptation.

\subsection{Generalized Pseudo-Label Driven Self-Training}
To bridge the visual domain gap driven by variable flight altitudes, sensor specifications, and GSD scaling, we implement an unsupervised self-training routine. 
This routine leverages the model optimized on the source domain to harvest latent spatial knowledge directly from the unannotated target imagery. 
The pipeline runs via a decoupled pseudo-label generation and joint-training loop. 
In the generation stage, target domain image patches from PotsRef are passed through the initialized source-domain model along with their corresponding text descriptions. 
The model outputs predicted probability maps, which we convert into hard binary masks using a fixed confidence threshold of 0.5. 
Our data loader incorporates an absolute control interface that maps low-confidence or non-reference target scenes to all-zero masks, preventing corrupted background noise from updating the segmentation head.
During the joint-training stage, we append the target image, the generated target pseudo-mask, and the target text description directly to the source training batches. 
The corresponding target text query is synthesized via the offline metadata lifting pipeline without exposing target masks to the optimization loop.
The data pipelines mix these streams into a unified training loop. The model minimizes a supervised objective using true ground truths for source samples and pseudo-labels for target samples, iteratively aligning the latent visual distributions without requiring manual target annotations.

\subsection{Multi-Granularity Text Prompt Mixing Strategy}
While self-training addresses visual distribution drift, it remains vulnerable to textual drift caused by unconstrained, variable user command granularities. 
To force the cross-modal alignment layer to build scale-invariant topological representations, we introduce a text prompt mixing strategy within the training loop.
Our data pipeline exploits the multi-tiered linguistic hierarchy established in the VPRef dataset. 
As implemented in the dataset loading classes, each target ground-truth mask is mapped to a synchronized pool containing Basic, Moderate, and Complex textual variations. 
During training, rather than fixing the text input to a single level of detail, we activate a dynamic mixing variant. 
For each training instance within an epoch, the text merger module randomly samples a single description from the multi-granularity text pool. The selection follows a uniform discrete distribution:
\begin{equation}
\mathcal{T}_{\text{input}} \sim {\mathcal{T}_{\text{Basic}}, \mathcal{T}_{\text{Moderate}}, \mathcal{T}_{\text{Complex}}}.
\end{equation}

This epoch-level resampling acts as a text-space data augmentation. By decoupling a fixed visual mask from a singular sentence structure, the model cannot rely on specific keyword patterns or fixed phrase lengths. 
This forces the model to accurately locate the target object, regardless of whether the description is short or long, simple or complex, successfully mitigating textual drift under domain transitions.

\subsection{Cross-Modal Matching and Joint Loss Optimization}
To optimize the network across both bounding box localization and dense pixel segmentation, we utilize a multi-task objective regulated by an assignment mechanism. 
Because our segmentation head generates a fixed set of prediction queries, we employ an optimized bipartite matching strategy \cite{carion2020end}. 
This matcher computes an optimal one-to-one assignment between the predicted queries and the target reference labels by minimizing a joint matching cost containing classification confidence, bounding box coordinates, and Generalized Intersection over Union (GIoU) metrics.
Once the optimal query assignments are established, the framework minimizes a composite joint loss function containing six distinct task-specific constraints:
\begin{equation}
\begin{aligned}
\mathcal L_{\text{total}}
&= \lambda_{\text{L1}} \mathcal L_{\text{L1}}
 + \lambda_{\text{GIoU}} \mathcal L_{\text{GIoU}}
 + \lambda_{\text{ce}} \mathcal L_{\text{ce}} \\
&\quad + \lambda_{\text{focal}} \mathcal L_{\text{focal}}
 + \lambda_{\text{Dice}} \mathcal L_{\text{Dice}}
 + \lambda_{\text{Presence}} \mathcal L_{\text{Presence}},
\end{aligned}
\end{equation}
where $\lambda_{\text{L1}}$, $\lambda_{\text{GIoU}}$, $\lambda_{\text{ce}}$, $\lambda_{\text{focal}}$, $\lambda_{\text{Dice}}$, and $\lambda_{\text{Presence}}$ are balancing coefficients, fixed at 2, 1, 5, 5, 5, and 2,
respectively.

The bounding-box localization terms comprise an L1 regression loss, $\mathcal L_{\text{L1}}$, on normalized box coordinates and a generalized intersection-over-union loss, $\mathcal L_{\text{GIoU}}$. 
The query-level classification term, $\mathcal L_{\text{ce}}$, follows SAM3's IoU-aware binary cross-entropy formulation, using IoU-dependent soft targets for matched positive predictions and confidence-dependent weighting for unmatched negative predictions. 
The dense segmentation terms comprise a binary focal loss, $\mathcal L_{\text{focal}}$, with $\alpha=0.25$ and $\gamma=2$, and a Dice loss, $\mathcal L_{\text{Dice}}$, providing pixelwise supervision and foreground-overlap supervision, respectively. 
The presence term, $\mathcal L_{\text{Presence}}$, supervises whether the image contains at least one valid target corresponding to the input description. 
Together, these terms provide supervision for bounding-box localization, query-level confidence, pixelwise segmentation, and image-level target presence during cross-domain adaptation.


\section{Experiments}
\label{experiments}

\subsection{Experimental Settings}
\label{subsec:experimental_settings}

\subsubsection{Evaluation Metrics}
To evaluate model performance in cross-domain RRSIS, we define two primary pixel-level mask metrics: mean Intersection over Union (mIoU) and overall Intersection over Union (oIoU). 
All results are reported in percentages.
Let $N$ denote the total number of test samples. For the $i$-th sample, let $P_i$ represent the predicted binary pixel mask and $G_i$ represent the corresponding ground-truth reference mask.
The mIoU metric computes the arithmetic mean of the intersection-over-union ratios calculated individually for each sample, assigning equal weight to each geographical target regardless of its pixel footprint:
\begin{equation}
\text{mIoU} = \frac{1}{N} \sum_{i=1}^{N} \frac{|P_i \cap G_i|}{|P_i \cup G_i|}.
\end{equation}

Concurrently, the oIoU metric aggregates the total intersecting pixels across the entire test distribution and divides them by the cumulative union area, effectively prioritizing large-scale geomorphic features such as extended industrial buildings or runways:

\begin{equation}
\text{oIoU} = \frac{\sum_{i=1}^{N} |P_i \cap G_i|}{\sum_{i=1}^{N} |P_i \cup G_i|}.
\end{equation}

\begin{table}[ht]
\centering
\caption{SAM3 Components Receiving LoRA Adapters.}
\label{exp:table0}
\begin{tabular}{lll}
\hline
SAM3 component	& \makecell{Adapted \\ linear layers}	& \makecell{Trainable \\ parameters}\\
\hline
Vision encoder (32 blocks)	&128	&4.52M \\
Text encoder (24 blocks)	&144	&3.54M\\
Geometry encoder (3 blocks)	&36	    &0.23M\\
DETR encoder (6 blocks)	    &60	    &0.42M\\
DETR decoder (6 blocks)   	&84	    &0.52M\\
Segmentation head	        &4	    &16,384\\
Dot-product scoring module	&2	    &8,192\\
Total	                    &458	&9.25M\\
\hline
\end{tabular}
\end{table}

\subsubsection{Baselines and Reference Models}
Our empirical framework assesses three prominent RRSIS architectures originally designed for single-domain scenarios: LAVT \cite{yang2022lavt}, LGCE \cite{yuan2024rrsis}, and RMSIN \cite{liu2024rotated}. 
To analyze the impact of large-scale parameters, we incorporate the SAM3 \cite{carion2026sam3segmentconcepts} under two distinct operating modes. 
First, we test the unmodified SAM3 configuration to establish an open-vocabulary reference baseline. 
Second, we integrate our low-rank, parameter-efficient adaptation layout to produce the fine-tuned baseline, SAM3-ft.

\subsubsection{Implementation Details}
We restrict adaptation gradients using a low-rank allocation with intrinsic rank $r=8$ and scaling coefficient $\alpha=16$, and apply a dropout rate of 0.5 to prevent feature co-adaptation \cite{srivastava2014dropout}. 
All pretrained SAM3 parameters, including biases, are frozen. The adapted model contains 9.25M trainable LoRA parameters across 458 linear layers, representing approximately 1.09\% of its total parameters.
Table \ref{exp:table0} lists the detailed SAM3 modules receiving LoRA adapters, together with the trainable parameter count for each component.
Optimization is conducted using the AdamW solver with an initial learning rate of $5 \times 10^{-5}$, balanced by a weight decay parameter of 0.01. 
Training executes across a uniform batch size of 16, distributed symmetrically across two NVIDIA RTX 5000 GPUs. For the conventional baseline networks, we preserve the optimization schedulers, learning rates, and tokenization steps native to their original publications. 
We apply data augmentation uniformly across all architectures, including random horizontal and vertical coordinate reflections. To neutralize baseline spatial grid mismatches, all source domain tiles from VaiRef and target domain tiles from PotsRef are dynamically resized to a standardized input resolution of $480 \times 480$ pixels during batch loading.

\begin{table*}[ht]
\centering
\caption{Quantitative Evaluation Under Textual Drift Across Granularity Tiers. Rows Indicate Training Tiers, While Columns Indicate Test Tiers.}
\label{exp:table2}
\small
\begin{tabular}{llrrrrrr}
\hline
Method & Train tier
& \multicolumn{2}{c}{Test: Basic}
& \multicolumn{2}{c}{Test: Moderate}
& \multicolumn{2}{c}{Test: Complex} \\
\cline{3-8}
& & mIoU & oIoU & mIoU & oIoU & mIoU & oIoU \\
\hline

LAVT \cite{yang2022lavt}
& Basic    & 78.81 & 73.08 & 40.81 & 43.05 & 40.04 & 41.30 \\
& Moderate & 65.25 & 61.26 & 78.68 & 73.20 & 76.03 & 71.28 \\
& Complex  & 73.32 & 68.43 & 78.32 & 72.68 & 77.66 & 72.31 \\
& Mix      & 78.16 & 72.35 & 78.27 & 72.38 & 77.31 & 71.90 \\
\hline

LGCE \cite{yuan2024rrsis}
& Basic    & 78.60 & 73.10 & 38.98 & 40.67 & 45.59 & 48.32 \\
& Moderate & 68.68 & 64.34 & 78.72 & 73.19 & 76.47 & 71.46 \\
& Complex  & 73.62 & 68.01 & 78.52 & 73.01 & 77.60 & 72.49 \\
& Mix      & 78.15 & 72.08 & 78.39 & 72.25 & 77.09 & 71.50 \\
\hline

RMSIN \cite{liu2024rotated}
& Basic    & 78.84 & 73.48 & 41.43 & 47.48 & 43.45 & 47.61 \\
& Moderate & 74.91 & 67.53 & 78.90 & 73.11 & 77.31 & 71.72 \\
& Complex  & 76.20 & 70.45 & 78.45 & 72.87 & 77.40 & 72.34 \\
& Mix      & 78.32 & 72.41 & 78.46 & 72.52 & 77.63 & 71.95 \\
\hline

SAM3-ft
& Basic    & 79.64 & 74.64 & 51.37 & 58.54 & 31.76 & 40.57 \\
& Moderate & 78.00 & 73.62 & 79.66 & 74.88 & 68.93 & 68.17 \\
& Complex  & 78.02 & 73.53 & 79.10 & 74.32 & 79.41 & 74.33 \\
& Mix      & 79.33 & 74.33 & 79.46 & 74.55 & 79.34 & 74.37 \\
\hline

SAM3 \cite{carion2026sam3segmentconcepts}
& Zero-shot & 47.38 & 50.35 & 29.43 & 34.87 & 25.92 & 29.52 \\
\hline
\end{tabular}
\end{table*}

\subsection{Main Quantitative Results and Cross-Domain Comparisons}
\label{subsec:main_results}

\subsubsection{Textual Granularity Shift and Robustness Analysis}

Table \ref{exp:table2} evaluates model robustness to textual drift within the source domain, VaiRef.
For each method, the rows specify the training tier, while the columns report performance on the Basic, Moderate, and Complex test tiers.
The matched-tier results correspond to entries where the training and test tiers are identical.
Under matched-tier evaluation, LAVT, LGCE, and RMSIN achieve mIoU scores ranging from 77.40\% to 78.90\%.
SAM3-ft achieves the highest matched-tier mIoU on all three tiers, with scores of 79.64\%, 79.66\%, and 79.41\% on Basic, Moderate, and Complex, respectively.
These matched-tier scores provide reference values for assessing performance under cross-tier evaluation.

The cross-tier entries reveal that performance degradation depends strongly on the direction of the tier transition.
Models trained on Basic descriptions exhibit substantial declines when evaluated on Moderate or Complex descriptions, whereas models trained on the richer description tiers generally show smaller declines when evaluated across tiers.
For example, SAM3-ft trained on Basic descriptions achieves 79.64\% mIoU on the Basic test tier, but only 51.37\% and 31.76\% on Moderate and Complex, respectively.
The Basic-to-Complex transition therefore produces an absolute decline of 47.88 percentage points.
These results indicate that training on Basic expressions alone is insufficient to maintain robust segmentation performance across the evaluated description tiers.

The Mix rows in Table \ref{exp:table2} report results obtained by randomly sampling a description from the three granularity tiers during training.
Across all four methods, the resulting mIoU scores remain within 0.65 percentage points of their corresponding matched-tier reference scores.
Specifically, SAM3-ft trained with mixed descriptions achieves 79.33\%, 79.46\%, and 79.34\% mIoU on the Basic, Moderate, and Complex test tiers, respectively.
These scores are only 0.31, 0.20, and 0.07 percentage points below the corresponding matched-tier results.
On the Complex test tier, mixed training improves SAM3-ft's mIoU from 31.76\% under Basic-only training to 79.34\%, an absolute gain of 47.58 percentage points.
These findings support multi-granularity prompt mixing as an effective strategy for maintaining segmentation performance across the evaluated description tiers.

\begin{table}[ht]
\caption{Quantitative Evaluation of Cross-Domain Generalization and Performance Gaps.}
\label{exp:table3}
\setlength\tabcolsep{4.2pt}
\begin{tabular}{clllllllll}
\hline
\multicolumn{1}{l}{} &  & \multicolumn{2}{c}{VaiRef test} &  & \multicolumn{2}{c}{PotsRef val} &  & \multicolumn{2}{c}{PotsRef test} \\
\multicolumn{1}{l}{} &  & mIoU & oIoU &  & mIoU & oIoU &  & mIoU & oIoU \\
\hline
\multirow{5}{*}{Basic} & LAVT  & 78.81 & 73.08 &  & 61.19 & 52.68 &  & 61.55 & 54.38 \\
 & LGCE & 78.6 & 73.1 &  & 60.26 & 53.34 &  & 61.26 & 55.54 \\
 & RMSIN & 78.84 & 73.48 &  & 60.21 & 52.94 &  & 60.96 & 55.2 \\
 & SAM3 & 47.38 & 50.35 &  & 43.9 & 48.15 &  & 45.43 & 49.38 \\
 & SAM3-ft & \textbf{79.64} & \textbf{74.64} &  & \textbf{70.45} & \textbf{65.64} &  & \textbf{71.53} & \textbf{67.72} \\
\hline
\multirow{5}{*}{Moderate} & LAVT  & 78.68 & 73.2 &  & 62.43 & 54.34 &  & 59.23 & 52.82 \\
 & LGCE & 78.72 & 73.19 &  & 60.53 & 52.98 &  & 58.99 & 52.71 \\
 & RMSIN & 78.9 & 73.11 &  & 60.93 & 52.31 &  & 59.07 & 51.33 \\
 & SAM3 & 29.43 & 34.87 &  & 30 & 36.34 &  & 28.26 & 34.77 \\
 & SAM3-ft & \textbf{79.66} & \textbf{74.88} &  & \textbf{70.48} & \textbf{65.57} &  & \textbf{65.78} & \textbf{64.56} \\
\hline
\multirow{5}{*}{Complex} & LAVT  & 77.66 & 72.31 &  & 60.3 & 51.16 &  & 59.17 & 52.35 \\
 & LGCE & 77.6 & 72.49 &  & 60.73 & 53.49 &  & 61.22 & 55.08 \\
 & RMSIN & 77.4 & 72.34 &  & 59.81 & 52.28 &  & 60.01 & 53.99 \\
 & SAM3 & 25.92 & 29.52 &  & 26.35 & 30.36 &  & 25.94 & 29.44 \\
 & SAM3-ft & \textbf{79.41} & \textbf{74.33} &  & \textbf{70.11} & \textbf{65.14} &  & \textbf{70.16} & \textbf{66.32} \\
\hline
\end{tabular}
\end{table}

\subsubsection{Visual Cross-Domain Generalization Gaps}
Table \ref{exp:table3} establishes a baseline diagnostic assessment of visual domain drift by evaluating source-trained models directly on the validation and testing partitions across all three text granularities. 
The empirical shifts quantify a pervasive performance gap across all evaluated architectures, revealing a severe generalization barrier when moving across geographic configurations.
On the cross-domain task, conventional models show a steep performance decline. Under the Basic text configuration, LAVT's mIoU drops 17.26 percentage points, from 78.81\% on the source test set to 61.55\% on the target test set. 
This vulnerability worsens in the Moderate tier, where RMSIN target test performance drops to 59.07\%, an absolute cross-domain gap of 19.83 percentage points in mIoU. 
The untuned SAM3 baseline fails across all cross-domain configurations because it cannot resolve fine-grained ground targets under shifting sensory footprints. 
Our SAM3-ft framework retains the highest absolute resilience to generalization across domains. SAM3-ft achieves 71.53\% mIoU on the PotsRef test set of the Basic tier. 
It still exhibits an absolute performance drop of 8.11 percentage points in mIoU relative to its source baseline. 

This uniform decline confirms that variations in visual features and spatial-scale discrepancies distort learned textual alignment coordinates, necessitating dedicated domain adaptation mechanisms.

\subsubsection{Unsupervised Domain Adaptation via Self-Training}

We perform a single round of offline pseudo-label generation and keep the pseudo-labels fixed throughout the 10 adaptation epochs.
With a fixed confidence threshold of 0.5, some pseudo-labels refer to null objects. 
To investigate their influence on self-training, we consider two variants. 
1) we keep these null-object pseudo-labels in the self-training round, termed \textit{-ssv};
2) we filter them out to create a cleaner pseudo-label variant \textit{-ssvf}.
For both types of self-training, the overall ratios of source and generated pseudo-labels are approximately 1:1.
Table \ref{exp:table4} isolates and quantifies the performance gains from integrating our pseudo-label-driven self-training framework across different model architectures. 
%
%
The empirical trends demonstrate that while self-training yields minor performance fluctuations on the source domain, it produces substantial and consistent improvements on the target domain. 
For example, on the PotsRef test set, the self-training plugin raises LAVT's baseline from 59.23\% to 60.89\% mIoU, while LAVT-ssvf reaches 63.09\% mIoU, an absolute gain of 3.86 percentage points in mIoU. 
Our parameter-efficient foundation model baseline capitalizes on this self-training loop. 
The SAM3-ft-ssv pipeline increases the PotsRef testing metric from 65.78\% to 67.27\% mIoU. 
By filtering out-of-distribution label noise, the framework achieves an absolute target peak of 67.46\% mIoU.


%


\begin{table}[]
\centering
\caption{Performance Gains of Unsupervised Self-Training Frameworks on the Moderate tier.}
\label{exp:table4}
\begin{tabular}{lllllllll}
\hline
& \multicolumn{2}{c}{VaiRef test}  & \multicolumn{2}{c}{PotsRef val}  & \multicolumn{2}{c}{PotsRef test} \\
 & mIoU & oIoU  & mIoU & oIoU   & mIoU & oIoU \\
\hline
LAVT  & \textbf{78.68} & \textbf{73.2}  & 62.43 & 54.34  & 59.23 & 52.82 \\
LAVT-ssv & 78.67 & 72.78   & \textbf{64.14} & \textbf{56.25}  & 60.89 & 54.5 \\
LAVT-ssvf & 78.57 & 72.6   & 63.78 & 56.24  & \textbf{63.09} & \textbf{56.81} \\
\hline
LGCE & 78.72 & \textbf{73.19}   & 60.53 & 52.98  & 58.99 & 52.71 \\
LGCE-ssv & 78.7 & 73.01   & 61.91 & \textbf{54.77}  & \textbf{63.1} & \textbf{56.58} \\
LGCE-ssvf & \textbf{78.75} & \textbf{73.19}  & \textbf{62.02} & 54.67  & 60.15 & 53.64 \\
\hline
RMSIN & \textbf{78.9} & \textbf{73.11}  & 60.93 & 52.31  & 59.07 & 51.33 \\
RMSIN-ssv & 78.8 & 72.86  & 61.96 & 53.75  & 61.03 & 54.02 \\
RMSIN-ssvf & 78.65 & 72.66  & \textbf{62.3} & \textbf{54.23}  & \textbf{63.54} & \textbf{56.26} \\
\hline
SAM3 & 29.43 & 34.87  & 30 & 36.34  & 28.26 & 34.77 \\
SAM3-ft & 79.66 & 74.88  & 70.48 & 65.57  & 65.78 & 64.56 \\
SAM3-ft-ssv & 79.55 & 74.63  & 70.85 & 66.11  & 67.27 & 65.54 \\
SAM3-ft-ssvf & \textbf{80.01} & \textbf{75.15}  & \textbf{71.15} & \textbf{66.22}  & \textbf{67.46} & \textbf{65.98} \\
\hline
\end{tabular}
\end{table}

\begin{table}[]
\caption{Ablation study of the Foundation Model Framework Integrating Multi-Granularity Text Mixing and Pseudo-Label Self-Training.}
\label{exp:table5}
\setlength\tabcolsep{5pt}
\begin{tabular}{llllllll}
\hline
 & \multicolumn{2}{c}{VaiRef test} & \multicolumn{2}{c}{PotsRef val}  & \multicolumn{2}{c}{PotsRef test} \\
 & mIoU & oIoU & mIoU & oIoU   & mIoU & oIoU \\
\hline
SAM3-ft & \underline{79.66} & \underline{74.88} &  70.48 & 65.57  & 65.78 & 64.56 \\
SAM3-ft-ssv & 79.55 & 74.63  & \underline{70.85} & 66.11  & 67.27 & 65.54 \\
SAM3-ft-ssvf & \textbf{80.01} & \textbf{75.15}  & \textbf{71.15} & \textbf{66.22}  & 67.46 & 65.98 \\
SAM3-ft-mix & 79.46 & 74.55 & 70.04 & 65.31   & \textbf{69.76} & \underline{66.45} \\
SAM3-ft-mix-ssv & 79.31 & 74.3 & 70.45 & 65.94   & \underline{68.54} & \textbf{66.59} \\
SAM3-ft-mix-ssvf & 79.52 & 74.45 & 70.41 & \underline{66.15}   & 68.08 & 65.95 \\
\hline
\end{tabular}
\end{table}

\subsubsection{Ablative Cross-Analysis of Foundation Model}
Table \ref{exp:table5} summarises the empirical results for the SAM3-ft baseline across various combinations of the multi-granularity text mixing strategy and pseudo-label-driven self-training. 
Rather than a single optimal configuration, the distribution of peak metrics across variants reveals metric-dependent effects of the two adaptation strategies.

%
The self-training variant SAM3-ft-ssv introduces minor feature corruption in the source dataset because uncalibrated out-of-distribution predictions propagate. 
With the pseudo-label filter, SAM3-ft-ssvf reverses this degradation, pushing the source partition to a maximum of 80.01\% mIoU and 75.15\% oIoU. 
This denoising mechanism also exhibits strong optimization stability on the target validation split.
%
%

Enriching the textual prompt space independently yields a profound regularizing effect on target-domain generalization. 
As shown by the SAM3-ft-mix results in Table \ref{exp:table5}, deploying the multi-granularity mixing strategy without self-training enables the framework to achieve the highest sample-level segmentation accuracy on the unseen target test partition, reaching an absolute peak of 69.76\% mIoU.
Because the dynamic mixing mechanism decouples a fixed spatial mask from a singular sentence length or syntactic template during training, it prevents the cross-modal transformer from exploiting localized linguistic shortcuts native to the source domain. 
The model is forced to distill invariant geometric and topological concepts, successfully insulating sample-level localization against textual drift.

When visual self-training couples with text-space randomization, the optimization trajectory shifts focus toward dominant geomorphic structures. 
The self-training variant, combined with text mixing (i.e., SAM3-ft-mix-ssv), achieves the highest cumulative pixel alignment on the target test partition, peaking at 66.59\% oIoU. 
Unfiltered self-training loops naturally gravitate toward majority land-cover classes, such as expansive impervious surfaces and building footprints, which command the largest pixel areas.
Simultaneously, the randomized text mixing layer acts as a data amplifier that robustifies the cross-modal boundaries of these dominant targets. This dual alignment forces the segmentation head to maximize overall pixel intersections for large-scale geographic elements, even if small-scale objects like vehicles experience marginal suppression. 
The resulting dispersion of peak metrics across \textit{-mix}, \textit{-ssvf}, and \textit{-mix-ssv} reflects generalization performance that depends on both the textual and the visual features, and indicates that combining the two strategies does not consistently outperform their individual use.

\begin{table}[]
\caption{Parameter Efficiency Analysis across Evaluated RRSIS Frameworks.}
\label{exp:table6}
\begin{tabular}{lllll}
\hline
 & \makecell[l]{Visual \\ Encoder}      & \makecell[l]{Text \\ Encoder}   & \makecell[l]{Total \\ parameters}  & \makecell[l]{Trainable \\ parameters}  \\
\hline
LAVT & Swin-B & BERT & 227.74M & 227.74M \\
LGCE & Swin-B & BERT & 276.26M & 276.26M \\
RMSIN & Swin-B & BERT & 240.04M & 240.04M \\
SAM3 & PE-L+ & PE-L+ & 849.76M & 0M \\
SAM3-ft & PE-L+ & PE-L+ & 849.76M & 9.25M \\
\hline
\end{tabular}
\end{table}

\subsubsection{Parameter Efficiency Analysis}
Table \ref{exp:table6} provides a diagnostic assessment of SAM3-ft on the VaiRef dataset, isolating the trade-offs between total capacity and parameter-efficient optimization scalability.

Conventional RRSIS models rely on identical sub-module combinations comprising a Swin-B vision backbone \cite{liu2021swin} coupled with a BERT text encoder \cite{devlin2019bert}. While these networks maintain a moderate memory footprint, with total parameter counts ranging from 227.74M to 276.26M, they leverage full-parameter training pathways. 
Full-weight backpropagation across these dense cross-modal pipelines not only demands intensive hardware execution budgets but also risks triggering catastrophic forgetting of foundational alignment priors due to unrestricted gradient contamination across model weights.

Concurrently, the unmodified foundational SAM3 dramatically expands the semantic ceiling by hosting a high-capacity PE \cite{bolya2026perception} for the vision and text encoders, scaling the collective parameter count to 849.76M. 
Although freezing the entire configuration yields a zero trainable footprint, this static operational mode remains decoupled from downstream remote sensing distributions, leaving its open-vocabulary spaces under-utilized without local spatial calibration. 
SAM3-ft successfully bridges this operational divide. By strategically embedding low-rank decomposition matrices within the foundational streams, SAM3-ft retains the massive 849.76M representational capacity while restricting active gradient pathways to a lightweight fraction of just 9.25M parameters. 
This design limits the trainable parameter footprint to approximately 1.09\% of the total model volume. Compared to conventional RRSIS baselines that require full-weight parameter backpropagation, SAM3-ft achieves superior segmentation performance while consuming a fraction of the trainable gradient budget. 
%

\begin{figure}[!t]
\centering
\includegraphics[width=3.5in]{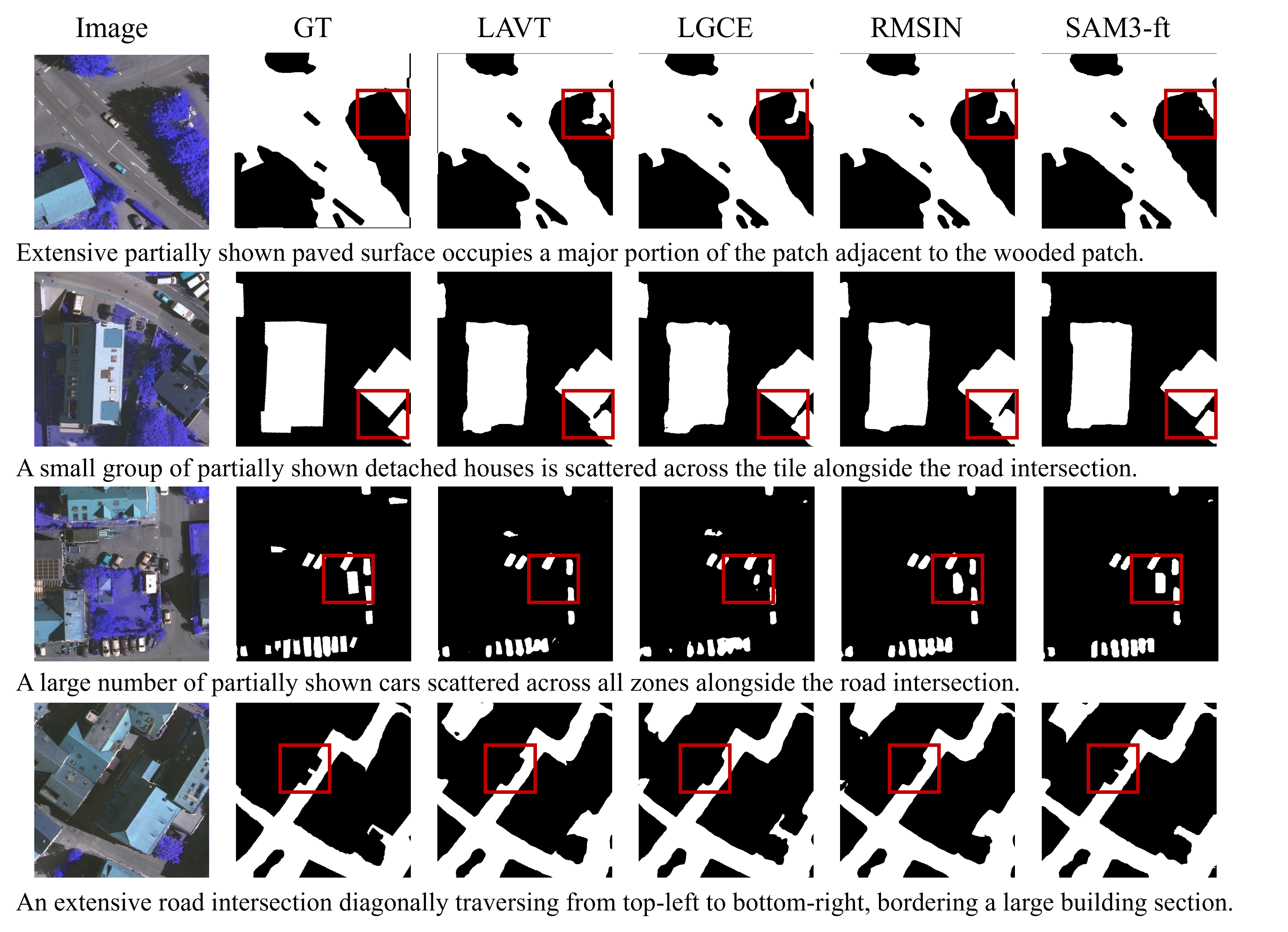}
\caption{Qualitative segmentation comparisons on the in-distribution VaiRef test set. The red bounding boxes highlight the superior multi-modal grounding capability of SAM3-ft in preserving continuous paved surface (row 1), crisp structural margins (row 2), and complete small-object details (rows 3--4) relative to full-parameter baselines.}
\label{result1}
\end{figure}

\begin{figure}[!t]
\centering
\includegraphics[width=3.5in]{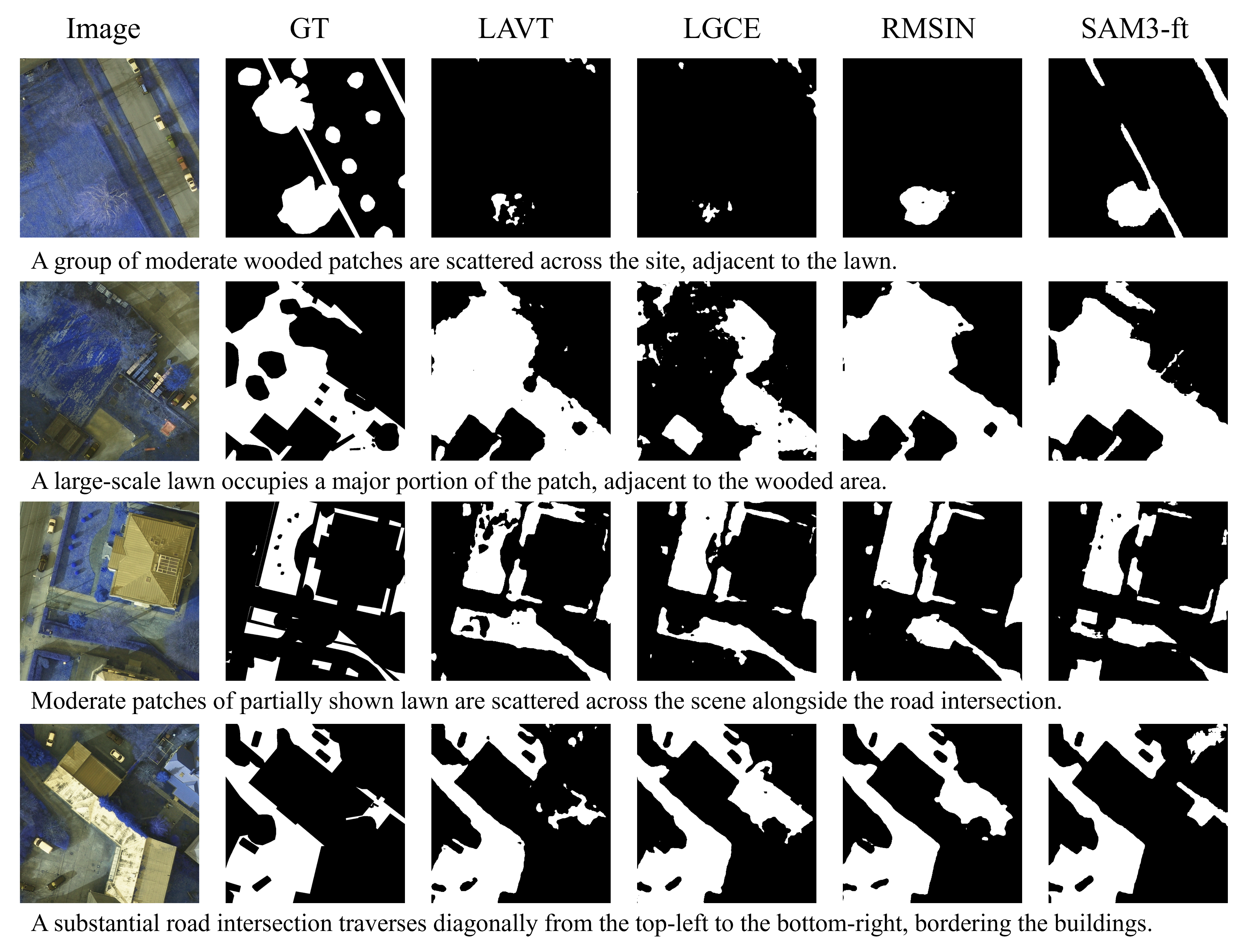}
\caption{Qualitative segmentation comparisons under cross-domain deployment on the PotsRef test set against spatial-spectral drift.}
\label{result2}
\end{figure}

\subsection{Qualitative Analysis and Visualizations}
\label{subsec:qualitative_analysis}

\subsubsection{Comparative Visual Evaluation Across RRSIS Baselines}

Fig. \ref{result1} and Fig. \ref{result2} present qualitative comparisons between our SAM3-ft baseline and conventional architectures across the VaiRef and PotsRef test sets. 
In the in-distribution VaiRef domain, the qualitative sequences in Fig. \ref{result1} confirm SAM3-ft's superior multimodal alignment capability across diverse land-cover classes. 
In the first row, highlighted by the red bounding box, SAM3-ft isolates the continuous, intact topology of the targeted paved street, whereas the masks from LAVT, LGCE, and RMSIN are visibly fragmented. 
For the residential structure in the second row, conventional RRSIS models yield blurred, imprecise boundaries, whereas SAM3-ft precisely traces the building's margins to match the ground truth. Furthermore, SAM3-ft resolves target instances with higher completeness in micro-scale contexts, such as the vehicles in the third row, and accurately distills intricate fine-grained details in complex mixed structures shown in the fourth row.

Under cross-domain conditions on the PotsRef test set in Fig. \ref{result2}, all networks show noticeable visual degradation due to mismatches in spatial resolution and spectral variations. Conventional frameworks display significant spatial misalignment, with LGCE failing to segment large, contiguous target regions. Symmetrically, LAVT and RMSIN suffer from severe target omissions and boundary erosion. In contrast, SAM3-ft exhibits the highest visual resilience, consistently generating the most stable, complete, and accurate mask structures among all evaluated models.

\subsubsection{Linguistic Granularity Analysis}
%
We evaluate the baseline SAM3-ft architecture on the VaiRef test set under an explicit cross-prompt granularity mismatch, keeping the model weights optimized on the Basic tier while systematically elevating the test prompts to Moderate and Complex descriptions.

\begin{figure*}[!t]
\centering
\includegraphics[width=7in]{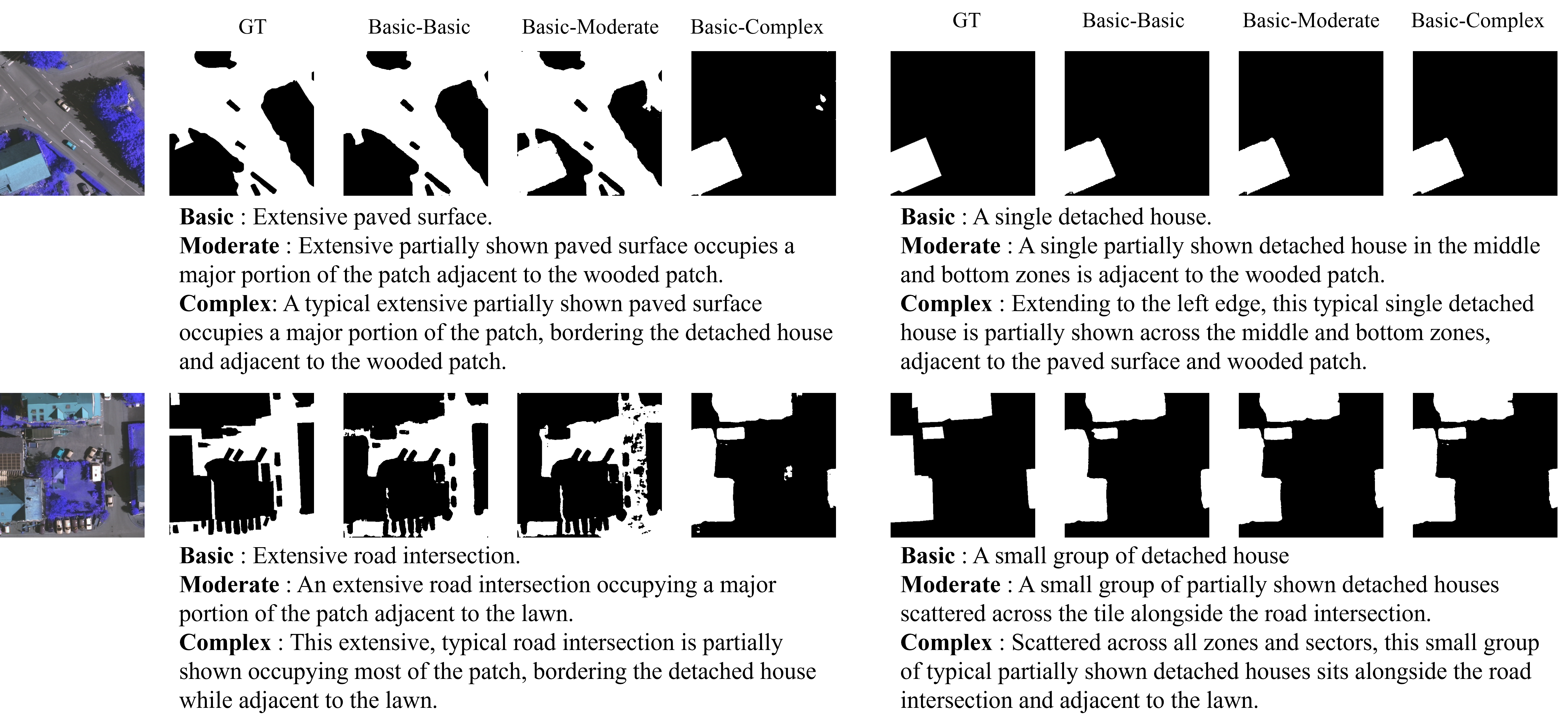}
\caption{Qualitative breakdown of cross-linguistic granularity inference on the VaiRef dataset at high levels of textual complexity.}
\label{result3}
\end{figure*}

As illustrated in Fig. \ref{result3}, this configuration reveals sensitivity to semantic interference and keyword distraction in long-tailed clauses. 
In the first row, where the target reference is an impervious paved surface, the model successfully segments the target under the Moderate prompt. However, under the Complex prompt, which introduces a long description with multiple secondary geomorphic objects, the cross-modal alignment mechanism becomes less effective. Instead of isolating the paved surface, the segmentation head completely shifts its focus and mistakenly segments the ``detached house" mentioned in the relative clause.

A parallel breakdown occurs in the second target sequence, which involves impervious surface extraction. When transitioning to the Complex textual query, the cross-modal transformer fails to maintain semantic hierarchy and misinterprets the structural description, incorrectly outputting the adjacent building structure instead of the road footprint.
This error pattern confirms that training an RRSIS model solely on brief keyword strings restricts its cross-modal attention field. Without dynamic text-space regularization during training, the network fails to construct an operational hierarchy for multi-layered sentences. 
%

\begin{figure}[!t]
\centering
\includegraphics[width=3.5in]{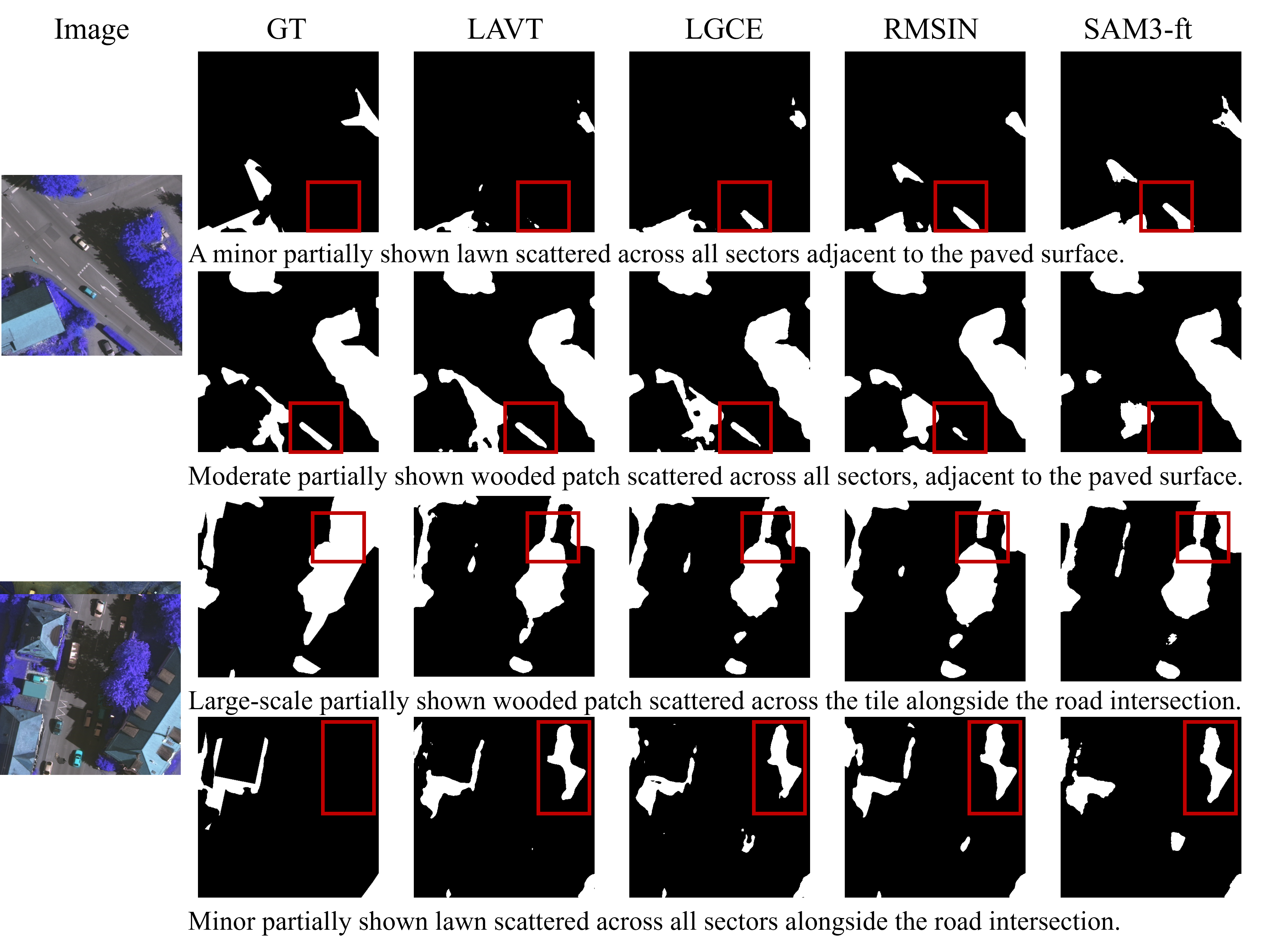}
\caption{Diagnostic error profiles and resilience to annotation-noise evaluations within the VaiRef test partition.}
\label{result4}
\end{figure}

\subsubsection{Error Profile Diagnostics and Annotation Noise Resilience}
Fig. \ref{result4} visually diagnoses the boundary conditions and inherent limits of pure spectral-textual cross-grounding, while highlighting our framework's resilience to dataset annotation noise. 
The qualitative sequences reveal distinct insights regarding elevation ambiguity and label corruption within the VaiRef test partition.

The first two rows illustrate a structural constraint that may stem from the lack of three-dimensional geometric data. When tasked with separating tall tree canopies from low vegetation variants under identical spectral zones, the cross-modal interaction layers exhibit localized height ambiguity. 
As highlighted by the red bounding box in the first row, the network misidentifies shrubbery as low-lying lawn because their multispectral signatures overlap heavily on the 2D grid. 
Symmetrically, in the second row, the model omits a lower-center tree cluster. 
These persistent discrepancies suggest that relying solely on two-dimensional appearance cues may limit the network's ability to resolve height-dependent vegetation tiers, indicating that incorporating auxiliary elevation models could be a plausible direction for structural disambiguation.

Conversely, the third and fourth rows show a compelling instance of resilience to annotation noise, where our fine-tuned baseline corrects errors in the manual ground-truth annotations. 
In the third row, the true reference target under deep shadow comprises shrubbery, yet the original dataset label shows potential annotation ambiguity and categorizes this zone as tree canopy.
Rather than blindly overfitting to this corrupted label distribution, these models suppress this localized label noise. 
In the fourth row, the baseline models correctly isolate the shrubbery boundaries and align perfectly with the structural query. 
This capacity to successfully bypass annotation artifacts demonstrates that the proposed framework does not merely memorize localized pixel-text associations, but instead distills transferable, structurally sound geomorphic concepts.

\section{Conclusion}
\label{conclusion}

In this paper, we introduce VPRef, the first cross-domain benchmark dataset for referring remote sensing image segmentation, specifically designed to assess model resilience to concurrent visual domain drift and textual drift. 
Symmetrically, we developed a generalized parameter-efficient domain adaptation baseline that integrates pseudo-label-driven self-training with a multi-granularity text prompt mixing strategy within a lightweight SAM3-LoRA optimization loop. 
Quantitative and qualitative cross-evaluations demonstrate that our framework successfully recovers from severe cross-domain performance drops, verifying that PEFT can stabilize multi-modal alignment coordinates under unknown target distributions. 
Despite these achievements, a key structural limitation of our current framework stems from the exclusive reliance on two-dimensional appearance cues; 
without auxiliary three-dimensional elevation information, the cross-modal interaction layers exhibit localized height ambiguity when separating tall tree canopies from spectrally overlapping low-lying vegetation. 
Future investigations will incorporate multi-source elevation data to resolve vertical geometric structures and explore source-free domain adaptation protocols to further enhance multi-modal deployment reliability in unconstrained Earth observation scenarios.




\bibliographystyle{IEEEtran}
\bibliography{reference}

\end{document}